%% file: main.tex
\documentclass[runningheads]{llncs}
\usepackage[]{graphicx}
\usepackage[]{color}
\usepackage{alltt}
\usepackage[utf8]{inputenc}
\usepackage{diagbox}
\usepackage{bbold}
\usepackage{makecell}
\usepackage{ragged2e}
\usepackage{xcolor}
\usepackage[nottoc]{tocbibind}
\usepackage{amsmath}
\usepackage{hyperref}
\usepackage{enumitem}

\newcommand{\proglang}[1]{\textsf{#1}}
\newcommand{\pkg}[1]{\textbf{#1}}

\usepackage[misc]{ifsym} 

\begin{document}

\title{Leveraging Model-based Trees as Interpretable Surrogate Models for Model Distillation}
%
\titlerunning{Leveraging Model-based Trees for Model Distillation}
%
\author{Julia Herbinger\thanks{These authors contributed equally to this work.}\inst{1,2}\orcidID{0000-0003-0430-8523} \and
Susanne Dandl$^*$\inst{1,2}\orcidID{0000-0003-4324-4163} \and
Fiona K. Ewald\inst{1,2}\orcidID{0009-0002-6372-3401} \and
Sofia Loibl\inst{1}\and
Giuseppe Casalicchio \Letter\,\inst{1,2}\orcidID{0000-0001-5324-5966}}
%
\authorrunning{J. Herbinger, S. Dandl et al.}
%
\institute{Department of Statistics, LMU Munich, Ludwigstr. 33, 80539 Munich, Germany \and
Munich Center for Machine Learning (MCML), Munich, Germany  \\
\email{Giuseppe.Casalicchio@stat.uni-muenchen.de}
}

\maketitle              
\begin{abstract}
Surrogate models play a crucial role in retrospectively interpreting complex and powerful black box machine learning models via model distillation. This paper focuses on using model-based trees as surrogate models which partition the feature space into interpretable regions via decision rules.
Within each region, interpretable models based on additive main effects are used to approximate the behavior of the black box model, striking for an optimal balance between interpretability and performance.
Four model-based tree algorithms, namely SLIM, GUIDE, MOB, and CTree, are compared regarding their ability to generate such surrogate models. We investigate fidelity, interpretability, stability, and the algorithms' capability to capture interaction effects through appropriate splits. 
Based on our comprehensive analyses, we finally provide an overview of user-specific recommendations.

\end{abstract}

\section{Introduction}
\label{intro}
\input{chapters/1_Introduction}

\section{Related Work}
\label{related}
\input{chapters/2_Related_work}

\section{Background: Model-Based Trees}
\label{background}
\input{chapters/3_Background}

\section{Comparison Study}
\label{simulation}
\input{chapters/5_Simulation}
\section{Discussion}
\label{conclusion}

\input{chapters/7_conclusion}

%
%
%

%



\newpage
\bibliographystyle{splncs04}
\bibliography{bibliography}
\end{document}


\title{Online Appendix for ``Leveraging Model-based Trees as Interpretable Surrogate Models for Model Distillation''}
%
\titlerunning{Leveraging Model-based Trees for Model Distillation}
%
\author{Julia Herbinger\thanks{These authors contributed equally to this work.}\inst{1,2}\orcidID{0000-0003-0430-8523} \and
Susanne Dandl$^*$\inst{1,2}\orcidID{0000-0003-4324-4163} \and
Fiona K. Ewald\inst{1,2}\orcidID{0009-0002-6372-3401} \and
Sofia Loibl\inst{1}\and
Giuseppe Casalicchio \Letter\,\inst{1,2}\orcidID{0000-0001-5324-5966}}
%
\authorrunning{J. Herbinger, S. Dandl et al.}

\institute{Department of Statistics, LMU Munich, Ludwigstr. 33, 80539 Munich, Germany \and
Munich Center for Machine Learning (MCML), Munich, Germany  \\
\email{Giuseppe.Casalicchio@stat.uni-muenchen.de}
}
%
%
\maketitle              
\appendix
\input{chapters/appendix}

\newpage
\bibliographystyle{splncs04}
\bibliography{bibliography}

%% file: chapters/1_Introduction.tex
Various machine learning (ML) algorithms achieve outstanding predictive performance, but often at the cost of being complex and not intrinsically interpretable.
This lack of transparency can impede their application, especially in highly regulated industries, such as banking or insurance \cite{Henckaerts.2022}.
A promising class of post-hoc interpretability methods to provide explanations for these black boxes are so-called surrogate models, which are intrinsically interpretable models -- such as linear models or decision trees  -- that approximate the predictions of black box models \cite{Molnar.2019}.
The learned parameters of surrogate models (e.g., the coefficients of a linear model or the tree structure) are thereby used to provide insights into the black box model.
The usefulness of these explanations hinges on how well they approximate the predictions of the original ML model. 
If a surrogate model is too simple to accurately approximate a complex black box model on the entire feature space, it cannot reliably explain the general behavior of the underlying ML model.
This is especially the case if the underlying ML model comprises feature interactions and highly non-linear feature effects. 
Other existing methods have, therefore, focused on local surrogate models to explain single observations. 
The idea is that while simple surrogate models 
may not accurately approximate the complex ML model on the entire feature space, they might be a good approximation in the immediate vicinity of a single observation. 
However, such local explanations cannot be used to explain the general model behavior.
We would require to produce and analyze multiple local explanations to get an understanding of the general model behavior, which is inconvenient 
because the sheer number of local models increases run-time while impeding interpretability. 
%

One promising idea to trade-off between global and local explanations is to train a global surrogate model that automatically finds regions in the feature space where the ML model’s predictions can be well described by interpretable surrogate models using only main effects.\footnote{In regions where only a few feature interactions are present, an additive main effect model is expected to be a good approximation.}
\cite{Hu.2020} introduced surrogate locally-interpretable models (SLIM) using model-based trees (MBTs) to find distinct and interpretable regions in the feature space where the ML model's predictions can be well described by a simple additive model that consists only of feature main effects in each leaf node. 
As such, SLIM generates regional additive main effect surrogate models which approximate the underlying ML model predictions and can be combined into a global surrogate model.
Other MBT algorithms have already been introduced before SLIM, but not as a surrogate model for post-hoc interpretation, for example, model-based recursive partitioning (MOB) \cite{Zeileis.2008}, conditional inference trees (CTree) \cite{Hothorn.2006}, and regression trees with unbiased feature selection and interaction detection (GUIDE) \cite{Loh.2002}. All these MBT algorithms are decision trees and usually differ in their splitting procedure and the objective used for splitting. 
While the well-known CART algorithm \cite{Breiman.1984} estimates constant values in the leaf nodes, MBTs fit interpretable models -- such as a linear model.

\paragraph{Contributions}
This paper aims to inspect SLIM, MOB, CTree, and GUIDE and their suitability as \textit{regional additive main effect surrogate models}\footnote{Meaning surrogate models that partition the feature space into interpretable regions and fit an additive main effect model within each region (e.g., a linear model using only first-order feature effects).} which approximate the underlying ML model predictions well. 
We specifically focus on main effect models because they enable good interpretability of the models in the leaf nodes. In the ideal case, interactions should then be handled by splits, so that the leaf node models are free from interaction effects.
In a simulation study, we apply the four MBT approaches as post-hoc surrogate models on the ML model predictions and compare them with regard to fidelity, interpretability, and stability -- as tree algorithms often suffer from poor stability \cite{Fokkema.2020}. 
We analyze their differences and provide recommendations that help users to choose the suitable modeling technique based on their underlying research question and data.



\paragraph{Reproducibility and Open Science}
The scripts to reproduce all experiments can be found at \url{https://github.com/slds-lmu/mbt_comparison}. 
It also contains the code for the SLIM and GUIDE algorithms, displaying (to the best of our knowledge) the first implementations of these approaches available in \proglang{R}. For the MOB and CTree algorithms the implementations of the \proglang{R} package \pkg{partykit} \cite{Hothorn.2015} were used.

%% file: chapters/2_Related_work.tex
The purpose of surrogate models is to approximate the predictions of a black box model as accurately as possible and to be interpretable at the same time  \cite{Molnar.2019}.
Global or local surrogate models are used based on whether the goal is to achieve a global interpretation of the black box model (model explanation) or to explain predictions of individual input instances (outcome explanation)
\cite{Maratea.2021}.


\paragraph{Global Surrogate Models}
The concept of global surrogate models is also known as model distillation, which involves training a simpler and more interpretable model (the distilled model) to mimic the predictions of a complex black box model. 
If the performance is good enough (i.e., high fidelity), the predictions of the black box model can be explained using the interpretable surrogate model. 
The main challenge is to use
an appropriate surrogate model that balances the trade-off between high performance 
and interpretability \cite{Molnar.2019}. For example, linear models are easily interpretable, but may not capture non-linear relationships modeled by the underlying ML model. 
Some researchers explore tree-based or rule-based approaches for model distillation \cite{bastani2017interpreting,frosst2017distilling,ming2018rulematrix}. 
Others propose promising models like generalized additive models plus interactions (GA2M) 
that include a small number of two-way interactions in addition to non-linear main effects to achieve both high performance and interpretability \cite{Lou.2013}.


\paragraph{Local Surrogate Models}
Local interpretable model-agnostic explanations (LIME) \cite{Ribeiro.2016} is probably the most prominent local surrogate model. It explains a single prediction by fitting a surrogate model in the local neighborhood around the instance of interest. This can, for example, be achieved by randomly sampling data points following the distribution of the training data and weighting the data points according to the distance to the instance of interest. 
This local approach offers an advantage over global surrogate models, as it allows for a better balance between model complexity and interpretability by focusing on a small region of the feature space, thereby achieving a higher fidelity in the considered locality.
However, the selection of an appropriate neighborhood for the instance of interest remains a challenging task \cite{Laugel.2018.limelocal}. 
A major drawback of LIME is its instability, as a single prediction can obtain different explanations due to different notions and possibilities to define the local neighborhood.
Several modifications of LIME have been proposed to stabilize the local explanation, including S-LIME \cite{zhou2021s} and OptiLIME \cite{visani2020optilime}.

\paragraph{Regional Surrogate Models}
The idea of regional surrogate models is to partition the feature space into appropriate regions in which fitting a simple interpretable model is sufficient.
For example, K-LIME \cite{Hall.2017} uses an unsupervised approach to obtain K partitions via K-means clustering. 
In contrast, locally interpretable models and effects based on supervised partitioning (LIME-SUP) \cite{Hu.2018} -- also known as SLIM \cite{Hu.2020} -- use a supervised approach (MBT) to partition the feature space according to a given objective. 

To the best of our knowledge, the suitability of well-established model-based tree algorithms (e.g., MOB, CTree, and GUIDE) as surrogate models has not been investigated so far.

%% file: chapters/3_Background.tex
In the following, the general framework underlying MBTs is introduced and the different algorithms are presented.

\subsection{General Notation}

We consider a $p-$dimensional feature space 
$\mathcal{X}$ 
and a target space $\mathcal{Y}$ (e.g., for regression $\mathcal{Y} = \mathbb{R}$ and for classification $\mathcal{Y}$ is finite and categorical with $|\mathcal{Y}| = g$ classes).
The respective random variables are denoted by $X = (X_1, \hdots, X_p)$ and $Y$.
The realizations of $n$ observations are denoted by $(y^{(1)}, \mathbf{x}^{(1)}),..., (y^{(n)}, \mathbf{x}^{(n)})$.
We further denote $\mathbf{x}_j$ as the $j$-th feature vector containing the observed feature values of $X_j$.
Following \cite{Zeileis.2008} and \cite{Seibold.2016}, let $\mathcal{M}((y, \mathbf{x}), \theta)$ be a parametric model, that describes the target $y$ as a function of a feature vector $\mathbf{x} \in \mathcal{X}$ and a vector of parameters $\mathbf{\theta} \in \Theta$. As a surrogate model the notation $\mathcal{M}((\hat{y}, \mathbf{x}), \theta)$ is used, i.e. the surrogate estimates the predictions of the black box model $\hat{y}$.
Thus, $y$ denotes the observed ground truth, $\hat{y}$ the black box model predictions, and $\hat{\hat{y}}$ the surrogate model predictions.
For regional surrogate models, the feature space is partitioned into $B$ distinct regions $\{\mathcal{B}_b\}_{b = 1,..., B}$ with the corresponding 
locally optimal vector of parameters $\theta_b$ in each partition $b = 1,..., B$.
    


\subsection{Model-based Tree (MBT) Algorithms}
In this section, the four MBT algorithms SLIM, MOB, CTree, and GUIDE are described, and theoretical differences are explained.
All MBTs can be described by the following recursive partitioning algorithm:
\begin{enumerate}
    \item Start with the root node containing all $n$ observations. 
    \item Fit the model $\mathcal{M}$ to all observations in the current node 
    to estimate $\hat{\theta}_b$.
    \item Find the optimal split within the node. 
    \item Split the current node into two child nodes until a certain stop criterion\footnote{For example,  until a certain depth of the tree, a certain improvement of the objective after splitting, or a certain significance level for the parameter instability is reached.} is met and repeat steps 2-4.
\end{enumerate}
SLIM uses an exhaustive search to select the optimal split feature and split point.
Due to the exhaustive search, SLIM might suffer from a selection bias similar to CART.\footnote{According to \cite{Hothorn.2006} an algorithm for recursive partitioning is called unbiased when, under the conditions of the null hypothesis of independence between target $y$ and features $\textbf{x}_{1}, ..., \textbf{x}_{p}$, the probability of selecting feature $\textbf{x}_{j}$ is $1/p$ for all $j = 1, ..., p$ regardless of the measurement scales or the number of missing values.} MOB, CTree, and GUIDE apply a 2-step procedure to combat the selection bias \cite{Schlosser.2019}:

\begin{enumerate}
    \item Select the feature with the highest association with the target $y$ to perform the splitting (partitioning). The hypothesis tests used to determine the most significant association differ between the MBT algorithms.
    \item Search for the best split point only within this feature (e.g., by exhaustive search or again by hypothesis testing).
\end{enumerate}
We will use all four algorithms as surrogate MBTs in conjunction with a linear main effect model 
$\mathcal{M}((\hat{y}, \mathbf{x}), \mathbf{\theta}_b) = \theta_{0,b} + \theta_{1,b} x_{1} + ... + \theta_{p, b} x_p$ with $\mathbf{\theta}_b = (\theta_{0,b}, ..., \theta_{p, b})^T$  for a (leaf) node $b$ such that the splits are based on feature interactions.  The assumption here is that if the main effects in the nodes are well-fitted, any lack of fit must come from interactions. Therefore, each feature can be used for splitting as well as for regressing (i.e., to train the linear main effect model in each node).
In the following, the four approaches are presented in more detail. Table \ref{tab:mbt_comparison} gives a concise comparison of them.

\begin{table}[t]
\centering
\setlength{\tabcolsep}{6pt} 
\caption{Comparison of MBT algorithms}
\label{tab:mbt_comparison}
\begin{tabular}{llll}
  \hline
 Algorithm& Split point selection & Test &  Implementation  \\ 
  \hline
    SLIM & exhaustive search & - &  - \\ 
    MOB & two-step & score-based fluctuation &  \proglang{R} \pkg{partykit} \\ 
    CTree & two-step & score-based permutation & \proglang{R} \pkg{partykit}\\ 
    GUIDE & two-step & residual-based $\chi^2$  &  binary executable\\ 
   \hline
\end{tabular}
\end{table}

\subsubsection{SLIM}

The SLIM algorithm performs an exhaustive search to find the optimal split point based on a user-defined objective function. \cite{Hu.2020} use the sum of squared errors -- similar to CART but fit more flexible parameterized models (such as an $L_1$-penalized linear model) instead of constant values. 
%
%
The computational effort for estimating all possible child models that are trained at each potential split point becomes very large with an increasing number of possible partitioning features.
For this reason, \cite{Hu.2020} developed an efficient algorithm for estimating them for the case of linear regression, linear regression with B-spline transformed features, and ridge regression (see \cite{Hu.2020} for more details).

To avoid overfitting and to obtain a small interpretable tree, \cite{Hu.2020} use the approach of post-pruning.
In order to keep the computational effort as low as possible, we use an early stopping mechanism: a split is only performed if the objective after the best split improves by at least a fraction of $\gamma \in [0,1]$ (compared to the objective in the parent node).
To the best of our knowledge, no openly accessible implementation of SLIM exists and we implemented SLIM in \proglang{R} as part of this work.

\subsubsection{MOB}

After an initial model is fitted in step 2, MOB examines whether the corresponding parameter estimates $\hat{\theta}_b$ are stable. 
To investigate this, the score function of the parametric model trained in the node is considered, which is defined by the gradient of the objective function with respect to the parameter vector $\theta_b$.
%
%
%
To test the null hypothesis of parameter stability, 
the M-fluctuation test is used \cite{Zeileis.2008}.
The feature for which the M-fluctuation test detects the highest instability (smallest p-value) is chosen for splitting. 
The choice of the optimal split point with respect to this feature is then made by means of an exhaustive search, analogous to SLIM.
MOB uses the Bonferroni-adjusted p-value of the M-fluctuation test as an early stopping criterion. That means a split is only performed if the instability is significant at a given significance level $\alpha$.
MOB generally distinguishes between regressor features, which are only used to fit the models in the nodes, and features, which are only used for splitting. However, \cite{Zeileis.2008} do not explicitly exclude overlapping roles as assumed here.
MOB is implemented in the $\pkg{partykit}$ \proglang{R} package \cite{Hothorn.2015}.


\subsubsection{CTree}
CTree -- similarly to MOB -- tries to detect parameter instability by measuring the dependency between potential splitting features and a transformation $h()$ of the target. 
A common transformation used in MBTs is the score function, which is also used for MOB. Also, other transformations such as the residuals could be used. 
However, \cite{Schlosser.2019} argues that the score function is generally preferred since it performs best in detecting parameter instabilities.

CTree uses a standardized linear association test to test the independence between the transformation $h()$ and the potential split features.
In the linear model case with continuous or categorical split features, this is equal to the Pearson correlation and one-way ANOVA test, respectively \cite{Schlosser.2019}.
%
The final test statistic follows an asymptotic $\chi^2$--distribution under the null hypothesis.
The feature with the smallest p-value is selected as the splitting feature. As for MOB, a Bonferroni-adjusted p-value is used as an early stopping criterion \cite{Hothorn.2006}.

Unlike SLIM and MOB, the split point is selected by a statistical hypothesis test. The discrepancy between two subsets is measured with a two-sample linear test statistic for each possible binary split. The split that maximizes the discrepancy is chosen as the split point \cite{Hothorn.2006}.
\cite{Schlosser.2019} state that the linear test used in CTree has higher power in detecting smooth relationships between the scores and the splitting features compared to the M-fluctuation test in MOB. MOB, on the other hand, has a higher ability in detecting abrupt changes.
An implementation of CTree is also part of the $\pkg{partykit}$ \proglang{R} package \cite{Hothorn.2015}.

\subsubsection{GUIDE}
GUIDE \cite{Loh.2002} uses a categorical association test to detect parameter instabilities. Specifically, a $\chi^2$--independence test between the dichotomized residuals at 0 (only the sign of the residuals matter) of the fitted model and the categorized features are performed and the p-values of these so-called curvature tests are calculated. In addition to the curvature tests, GUIDE explicitly searches for interactions by using again $\chi^2$--independence tests. 
If the smallest p-value comes from a curvature test, the corresponding feature is chosen as the partitioning feature. If the smallest p-value is from an interaction test, the categorical feature involved, if any, is preferably chosen as the splitting feature. If both involved features are categorical, the feature with the smaller p-value of the curvature test is chosen for splitting. In the case of two numerical features, the choice is made by evaluating the potential child models after splitting with respect to both features.
Subsequently, a bootstrap selection bias correction is performed.
In the original GUIDE algorithm developed by \cite{Loh.2002}, categorical features are only used for splitting due to the large number of degrees of freedom that are needed for the parameter estimation of categorical features.
GUIDE is only available as a binary executable under \url{https://pages.stat.wisc.edu/~loh/guide.html}.
We incorporated GUIDE as an option for the SLIM implementation in R. Pruning is therefore carried out in the same way as for SLIM.

%% file: chapters/5_Simulation.tex
Here, we first define desirable properties of MBTs when used as surrogate models on the predictions of an ML model. To quantify these properties and compare them for the different MBT algorithms, we define several evaluation measures. Then, we compare SLIM, MOB, CTree, and GUIDE based on these measures for different experimental settings and provide recommendations for the user.

\subsection{Desirable Properties and Evaluation Measures}
The following properties of MBTs are desirable:

\begin{itemize}
    \item \textbf{Fidelity}: To derive meaningful interpretations, the predictions of the ML model need to be well approximated by the MBT.
    \item \textbf{Interpretability}: To provide insights into the inner workings of the ML model, the MBT needs to be interpretable and hence not too complex to be understood by a human.
    \item \textbf{Stability}: Since MBTs are based on decision trees, they might be unstable in the sense that they are not robust to small changes in the training data \cite{Fokkema.2020}. However, stable results are needed for reliable interpretations.
\end{itemize}
These properties are measured using the evaluation metrics below.

\subsubsection{Fidelity}
To evaluate the fidelity of an MBT as a surrogate model, we use the  $R^2$, which is defined by $$\textstyle R^2\left( \left\{\hat{y}, \hat{\hat{y}}\right\}\right) = 1-\frac{\sum_{i = 1}^{n}\left(\hat{\hat{y}}^{(i) } - \bar{\hat{y}}\right)^2}{\sum_{i = 1}^{n}\left(\hat{y}^{(i)} - \bar{\hat{y}}\right)^2},$$
where $\hat{\hat{y}}^{(i)}, i = 1,...,n$ are the predictions of the MBT model and $\bar{\hat{y}}$ is the arithmetic mean of the ML model predictions $\hat{y}^{(i)}$.
%
%
Fidelity is measured on training data but also on test data in order to evaluate the MBT's fidelity on unseen data.

\subsubsection{Interpretability}
If different MBTs fit the same interpretable models within the leaf nodes (as done in the following experiments), the interpretability comparison of MBTs reduces to the complexity of the respective tree structure.
Therefore, the number of leaf nodes is used here to evaluate the interpretability, since fewer leaf nodes lead to shallower trees which are easier to understand and interpret. 


\subsubsection{Stability}
We consider an MBT algorithm stable if it partitions the feature space in the same way after it has been applied again on slightly different training data. To compare two MBTs, an additional evaluation data set is used that is partitioned according to the decision rules learned by each of the two MBTs. 
If the partitioning is identical for both MBTs, the interpretation of the decision rules is also assumed to be identical, which suggests stability. 

To measure the similarity of regions found by MBTs trained on slightly different data, the Rand index (RI) \cite{Rand.1971}, which was introduced for clustering approaches, is used. The RI defines the similarity of two clusterings $\mathcal{A}, \mathcal{B}$ of $n$ observations each by the proportion of the number of observation pairs that are either assigned to the same partition in both clusterings ($n_{11}$) or to different partitions in both clusterings ($n_{00}$) measured against the total number of observation pairs \cite{Gates.2017}:
$$\text{RI}(\mathcal{A}, \mathcal{B}) = \frac{n_{11} + n_{00}}{\binom{n}{2}}.$$

When comparing MBTs, the clusterings in the RI are defined by the regions based on the decision rules learned by the MBTs. Since the number of leaf nodes influences the RI, we only compute RIs for MBTs with the same number of leaf nodes.
A high RI for a pair of MBTs with the same number of leaf nodes indicates that the underlying MBT algorithm is stable for the analyzed scenario.

Additionally, the range of the number of leaf nodes is used as a measure of stability. It is assumed that MBTs are more unstable if the number of leaf nodes for different simulation runs varies strongly. 


\subsection{Experiments}
\label{sec:experiments}
Here, we empirically evaluate the four presented MBT algorithms with respect to fidelity, interpretability, and stability as surrogate models to interpret the underlying ML model. Therefore, we define three simple scenarios (linear smooth, linear categorical, linear mixed) which mainly differ regarding the type of interactions. Thus, we evaluate how well the MBT algorithms can handle these different types of interactions to provide recommendations for the user at the end of this section, depending on the underlying data and research question.

\subsubsection{Simulation setting}
Since the measures for interpretability and fidelity strongly depend on the early stopping configuration of $\gamma$ for SLIM and GUIDE and $\alpha$ for MOB and CTree, the simulations are carried out for three different values of each of these parameters for all three scenarios and for a sample size of $n = 1500$ of which $n_{train} = 1000$ observations are used for training and $n_{test} = 500$ observations for testing.
All MBT algorithms are fitted as surrogate models on the predictions of a correctly specified linear model (lm) or generalized additive model (GAM) and on an XGBoost model with correctly specified interactions. Further specifications of the hyperparameters for the XGBoost algorithm for each scenario can be found in Online~Appendix~\ref*{app:config}~\cite{HLDEC23}.
Table \ref{tab:simulation_setting} provides an overview of all $3 \times 3 \times 2 = 18$ variants for each of the four MBT algorithms.
Hyperparameters that are fixed in all variants are a  maximum tree depth of $6$ and a minimum number of observations per node of $50$.
We perform $100$ repetitions.
\begin{table}[!htb]
\centering 
\caption{Definition of variants for all simulation settings.}
\label{tab:simulation_setting}
\begin{tabular}[t]{lll}
\hline
Varied factors & levels \\
\hline
Scenario  & linear smooth, linear categorical, linear mixed\\
Early stopping config.   & $\alpha \in \{0.05,0.01,0.001\}, \gamma \in \{0.05,0.1,0.15\}$ \\
Surrogate model  & lm/GAM, XGBoost \\
\hline
\end{tabular}
\end{table}
Fidelity and interpretation measures are calculated in each simulation run.
The RIs are calculated after the simulation based on pairwise comparisons of the final regions of an evaluation data set. More detailed steps on the quantification of the RI for the stated simulation settings are explained in Online~Appendix~\ref*{app:sim_setup}~\cite{HLDEC23}.


\subsubsection{Linear smooth}
\label{sec:lin_smooth}

\paragraph{Scenario definition}
The DGP in this scenario includes one smooth two-way interaction between two numeric features and is defined as follows:
Let $X_1, X_2, X_3 \sim \mathcal{U}(-1,1)$, then the DGP based on the $n$ drawn realizations is given by $y = f(\textbf{x}) + \epsilon$ with
$f(\textbf{x}) = \textbf{x}_1 + 4   \textbf{x}_2 + 3   \textbf{x}_2   \textbf{x}_3 $ and
$\epsilon \sim \mathcal{N}(0, 0.01 \cdot \sigma^2(f(\textbf{x})))$.

\paragraph{Results}

Aggregated results on interpretability and fidelity are provided for all four MBT algorithms as surrogate models on the respective black box model in Table \ref{tab:linear_smooth_summary}.
Since the DGP is rather simple, all MBTs have a high fidelity but they also require a very high number of leaf nodes since the smooth interactions can only be well approximated by many binary splits.
To compare the fidelity, we focus on configuration $\gamma = \alpha = 0.05$ since this configuration leads to a similar mean number of leaf nodes for all four algorithms.
We see that, for a similar number of leaf nodes, the fidelity is equally high for the four MBTs, whereby the fidelity of MOB and CTree deviate less over the repetitions.

The number of leaf nodes fluctuates considerably more for SLIM and GUIDE than for MOB and CTree even when $\gamma$ and $\alpha$ are fixed.
In general, we can see that the $R^2$, which measures fidelity, increases with an increasing number of leaf nodes for all models, reflecting the trade-off between fidelity and interpretability.

\begin{table}[!htb]
\centering
\caption{Simulation results on 100 simulation runs for all four MBTs as surrogate models on the scenario linear smooth with sample sizes $n_{train}=1000$ and $n_{test} = 500$ for different values of $\gamma$ and $\alpha$. The mean (standard deviation) fidelity on the training data for the lm is 0.9902 (0.0006) and for the XGBoost 0.9858 (0.0008). On the test data set the respective fidelity values for the lm are 0.9901 (0.0008) and for the XGBoost 0.9768 (0.0018).}
\label{tab:linear_smooth_summary}
\begin{tabular}[t]{l|l|r|r|r|r|r|r|r|r}
\hline
Black box& MBT&$\gamma$/$\alpha$ & \multicolumn{3}{c|}{Number of Leaves}&\multicolumn{2}{c|}{$R^2_{train}$}&\multicolumn{2}{c}{$R^2_{test}$}\\
\hline
& &  & mean & min & max & mean & sd & mean & sd \\
\hline
lm & SLIM & 0.15 & 2.06 & 2 & 3 & 0.9650 & 0.0043 & 0.9631 & 0.0046\\
lm & SLIM & 0.10 & 12.11 & 5 & 16 & 0.9965 & 0.0052 & 0.9958 & 0.0060\\
lm & SLIM & 0.05 & 15.70 & 14 & 16 & 0.9995 & 0.0001 & 0.9993 & 0.0001\\
lm & GUIDE & 0.15 & 2.07 & 2 & 3 & 0.9651 & 0.0044 & 0.9632 & 0.0049\\
lm & GUIDE & 0.10 & 12.03 & 5 & 16 & 0.9965 & 0.0051 & 0.9957 & 0.0060\\
lm & GUIDE & 0.05 & 15.75 & 14 & 16 & 0.9995 & 0.0001 & 0.9993 & 0.0001\\
lm & MOB & 0.001 & 15.78 & 14 & 16 & 0.9994 & 0.0001 & 0.9993 & 0.0001\\
lm & MOB & 0.010 & 15.78 & 14 & 16 & 0.9994 & 0.0001 & 0.9993 & 0.0001\\
lm & MOB & 0.050 & 15.78 & 14 & 16 & 0.9994 & 0.0001 & 0.9993 & 0.0001\\
lm & CTree & 0.001 & 15.22 & 13 & 17 & 0.9993 & 0.0001 & 0.9992 & 0.0001\\
lm & CTree & 0.010 & 15.22 & 13 & 17 & 0.9993 & 0.0001 & 0.9992 & 0.0001\\
lm & CTree & 0.050 & 15.22 & 13 & 17 & 0.9993 & 0.0001 & 0.9992 & 0.0001\\
\hline
XGBoost & SLIM & 0.15 & 2.31 & 2 & 6 & 0.9665 & 0.0069 & 0.9629 & 0.0079\\
XGBoost & SLIM & 0.10 & 7.33 & 2 & 14 & 0.9850 & 0.0060 & 0.9814 & 0.0062\\
XGBoost & SLIM & 0.05 & 14.30 & 8 & 17 & 0.9948 & 0.0010 & 0.9909 & 0.0017\\
XGBoost & GUIDE & 0.15 & 2.26 & 2 & 5 & 0.9664 & 0.0067 & 0.9628 & 0.0077\\
XGBoost & GUIDE & 0.10 & 6.92 & 2 & 14 & 0.9847 & 0.0061 & 0.9811 & 0.0062\\
XGBoost & GUIDE & 0.05 & 14.15 & 8 & 17 & 0.9945 & 0.0010 & 0.9906 & 0.0017\\
XGBoost & MOB & 0.001 & 10.89 & 8 & 13 & 0.9944 & 0.0005 & 0.9904 & 0.0011\\
XGBoost & MOB & 0.010 & 11.96 & 9 & 15 & 0.9946 & 0.0005 & 0.9906 & 0.0011\\
XGBoost & MOB & 0.050 & 12.86 & 11 & 15 & 0.9948 & 0.0005 & 0.9908 & 0.0011\\
XGBoost & CTree & 0.001 & 12.09 & 9 & 15 & 0.9940 & 0.0006 & 0.9900 & 0.0012\\
XGBoost & CTree & 0.010 & 13.21 & 10 & 15 & 0.9943 & 0.0006 & 0.9902 & 0.0013\\
XGBoost & CTree & 0.050 & 14.09 & 11 & 17 & 0.9944 & 0.0006 & 0.9904 & 0.0012\\
\hline
\end{tabular}

\end{table}

\begin{figure}[!htb]
\centering
    \centering
    \includegraphics[width=\textwidth]{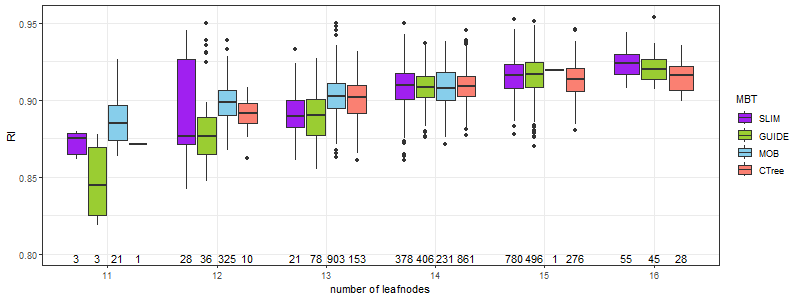}
    \caption{RI for the four MBT algorithms when used as surrogate models for an XGBoost model for the scenario linear smooth with $n_{train}=1000$ and $n_{test} = 500$, $\alpha = \gamma = 0.05$. The numbers below the boxplots indicate the number of tree pairs, for which both trees have the respective number of leaf nodes. Higher values are better.}
    \label{fig:ls_1000_xgboost_r2_train}
\end{figure}

Considering stability, SLIM and GUIDE provide -- for each configuration of $\gamma$ -- similar results for the number of leave nodes (interpretability) and for the $R^2$ values (fidelity). 
While these measures are rather sensitive with regard to the value of $\gamma$, the variation of $\alpha$ has a much smaller impact on the results for MOB and CTree.
Moreover, Figure \ref{fig:ls_1000_xgboost_r2_train} shows the RIs of the four algorithms applied to the XGBoost model for tree pairs with identical numbers of leaf nodes.
For a lower number of leaf nodes, MOB and CTree seem to generate more stable trees for this scenario. This effect diminishes with an increasing number of nodes and is also not apparent for the linear model (see Figure~1 in Online~Appendix~\ref*{app:lin_smooth}~\cite{HLDEC23}).

\subsubsection{Linear categorical}

\paragraph{Scenario definition}
Here, a scenario definition based on \cite{Herbinger.2022} with linear main effects and interactions between numerical and binary features (i.e., subgroup-specific linear effects) is regarded:
Let
$X_1, X_2 \sim \mathcal{U}(-1,1)$, $X_3 \sim Bern(0.5)$, then the DGP based on $n$ drawn realizations is defined by $y = f(\textbf{x}) + \epsilon$ with
$f(\textbf{x}) =  \textbf{x}_{1} - 8  \textbf{x}_2 + 16  \textbf{x}_2  \mathbb{1}_{(\textbf{x}_3 = 0)} + 8  \textbf{x}_2  \mathbb{1}_{(\textbf{x}_1 > \bar{\textbf{x}_1})} $ and
$\epsilon \sim \mathcal{N}(0, 0.01 \cdot \sigma^2(f(\textbf{x})))$.
In this scenario, the DGP is not determined by a smooth interaction but can be fully described by main effect models in four regions.
Assuming that the ML model accurately approximates the real-world relationships, if the regions of the MBTs are defined by a (first-level) split with respect to the binary feature $\textbf{x}_3$ and by a (second-level) split at the empirical mean of feature $\textbf{x}_1$, the final regions only contain main effects of the given features as defined in the DGP.

\paragraph{Results}

Aggregated results on interpretability and fidelity are provided for all four MBT algorithms as surrogate models on the respective black box model 
in Online~Appendix~\ref*{app:lin_cat}~\cite{HLDEC23}.
In all scenarios -- independent of the early stopping configurations -- SLIM and GUIDE lead to fewer leaves than MOB and CTree. This can be explained by the chosen split features.
SLIM and GUIDE perform splits with respect to features $\textbf{x}_1$ and $\textbf{x}_3$, which lead to the subgroup-specific linear effects defined by the DGP and thus, only need a few splits to well approximate the DGP. In contrast, MOB and CTree rather choose $\textbf{x}_2$ for splitting, resulting in many splits to achieve a comparable fidelity.
Hence, MOB and CTree lead to a worse fidelity and due to many more leaf nodes to a worse interpretability than SLIM and GUIDE for this scenario.

\begin{figure}[t]
\centering
    \includegraphics[width=\textwidth]{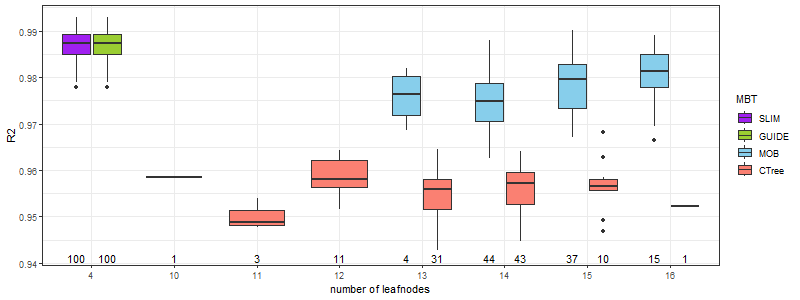}
    \caption{Test accuracy $R^2$ vs. number of leaf nodes for the four MBT algorithms as surrogate models for XGBoost for scenario linear categorical with $n_{train}=1000$ and $n_{test} = 500$, $\alpha = \gamma = 0.05$. The numbers below the boxplots indicate the number of trees (from 100 simulation runs) which have the respective number of leaf nodes for the regarded algorithm.}
    \label{fig:la_1000_standalone_r2_test}

\end{figure}

Figure \ref{fig:la_1000_standalone_r2_test} shows that, while the number of leaf nodes for SLIM and GUIDE is always four for the regarded setting ($\alpha = \gamma = 0.05$, see Online~Appendix~\ref*{app:lin_cat}~\cite{HLDEC23} for an overview), the number of leaf nodes for MOB and CTree varies strongly for the different simulation runs.
It is also noticeable that MOB performs better than CTree with the same number of leaf nodes.
A possible explanation is that the fluctuation test used within the splitting procedure of MOB performs better in detecting abrupt changes than the linear test statistic used in CTree \cite{Schlosser.2019}.
%
%
Online~Appendix~\ref*{app:lin_cat}~\cite{HLDEC23} also shows that SLIM and GUIDE show a better mean fidelity compared to MOB and CTree when applied to a GAM.

\subsubsection{Linear mixed}
\paragraph{Scenario definition}
The third scenario combines the linear smooth and the linear categorical scenarios. Hence, the DGP is defined by
linear main effects, interaction effects between categorical and numerical features, and smooth interactions:
Let $X_1, X_2 \sim \mathcal{U}(-1,1)$, $X_3, X_4 \sim Bern(0.5)$, then the DGP based on $n$ drawn realizations is defined by $y = f(\textbf{x}) + \epsilon$ with
$f(\textbf{x}) = 4   \textbf{x}_2 + 2   \textbf{x}_4  + 4   \textbf{x}_2   \textbf{x}_1 + 8   \textbf{x}_2   \mathbb{1}_{(\textbf{x}_3 = 0)} +  8 \textbf{x}_1   \textbf{x}_2 \mathbb{1}_{(\textbf{x}_4 = 1)}$ and
$\epsilon \sim \mathcal{N}(0, 0.01 \cdot \sigma^2(f(x)))$.

\paragraph{Results}

\begin{figure}[!htb]
\centering
    \includegraphics[width=\textwidth]{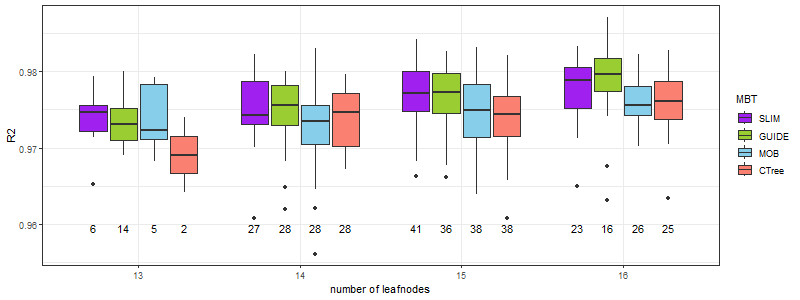}
    \caption{Test accuracy $R^2$ vs. number of leaf nodes for the four MBT algorithms when used as a surrogate model on the xgboost model for scenario linear mixed with $n_{train}=1000$ and $n_{test} = 500$, $\alpha = \gamma = 0.05$. The numbers below the boxplots indicate the number of trees (from 100 simulation runs) which have the respective number of leaf nodes for the regarded algorithm.}
    \label{fig:lm_1000_standalone_r2_test}

\end{figure}

\begin{figure}[!htb]
\centering
    \includegraphics[width=\textwidth]{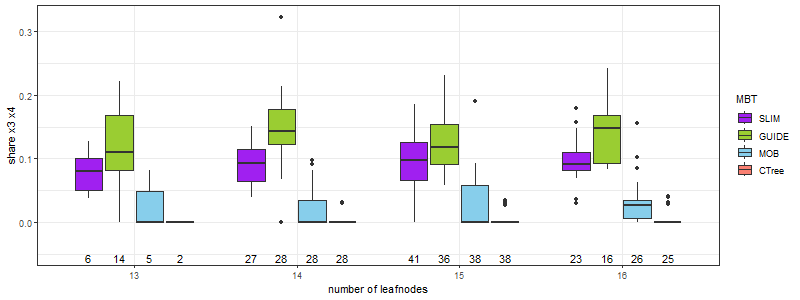}
    \caption{Relative amount of partitions which use features $\textbf{x}_3, \textbf{x}_4$ for splittings vs. the number of leaf nodes for the four MBT algorithms when the XGBoost model is used as a surrogate model for the scenario linear mixed with $n=1500, \alpha = \gamma = 0.05$.}
    \label{fig:lm_1000_standalone_share_x3x4}

\end{figure}

Aggregated results on interpretability and fidelity are provided for all four MBT algorithms as surrogate models on the respective black box model 
in Online~Appendix~\ref*{app:lin_mixed}~\cite{HLDEC23}.
To compare the different MBT algorithms, we choose the early stopping configurations, which lead to a similar mean number of leaf nodes ($\gamma = \alpha = 0.05$).
Figure \ref{fig:lm_1000_standalone_r2_test} shows that SLIM and GUIDE achieve a slightly better trade-off between fidelity and interpretability than MOB and CTree in this scenario.
This can be reasoned as follows:
SLIM and GUIDE split more often with respect to the binary features compared to the other two MBT algorithms (see Figure \ref{fig:lm_1000_standalone_share_x3x4}). Thus, SLIM and GUIDE use the categorical features more often to reveal the subgroups defined by them, while MOB and particularly CTree split almost exclusively with respect to the numerical features and hence perform slightly worse for the same mean number of leaf nodes.

%
%

\begin{figure}[!htb]
    \includegraphics[width=\textwidth]{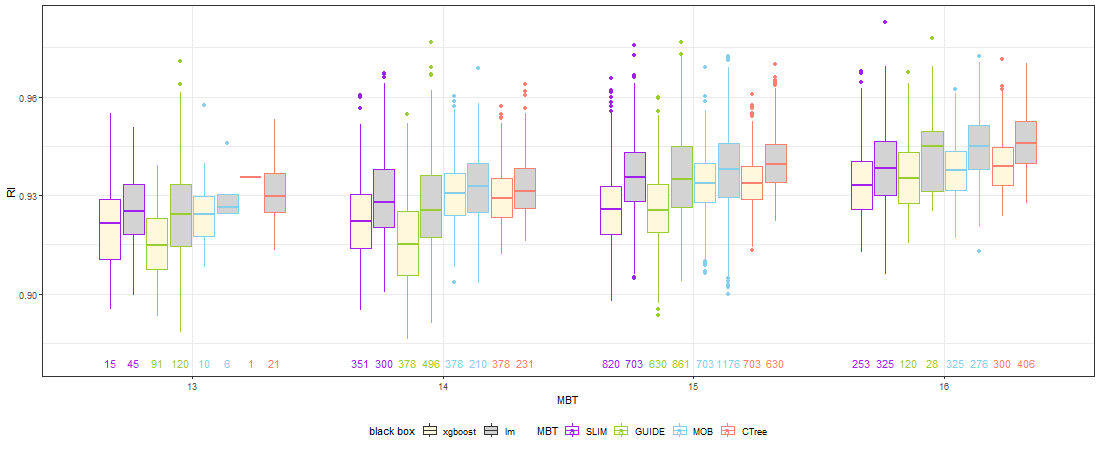}
    \caption{RI for the four MBT algorithms when used as a surrogate model for the scenario linear mixed with $n_{train}=1000$ and $n_{test} = 500$, $\alpha = \gamma = 0.05$. The numbers below the boxplots indicate the number of tree pairs (from 4950 pairs), for which both trees have the respective number of leaf nodes. Higher values are better.}
    \label{fig:lm_1000_standalone_lm_sta}
\end{figure}

However, MOB and CTree provide on average more stable results regarding the RI for a fixed number of leaf nodes compared to SLIM and GUIDE (see Figure \ref{fig:lm_1000_standalone_lm_sta}).

\subsection{Recommendations}
Based on our analyses in this work, we provide some general recommendations on how to choose the MBT algorithm based on the underlying data and research question. The recommendations are based on the given assumption that we are interested in partitioning the feature space in such a way that we receive interpretable and distinct regions where regional relationships are reduced to additive (linear) main effects of the features. Hence, the feature space is partitioned such that feature interactions are reduced.

If features cannot be separated into modeling and partitioning features (as is the case here), we recommend to
\begin{enumerate}
    \item use SLIM and GUIDE on subgroup detection tasks (scenario linear categorical and linear mixed) since they provide a better trade-off between fidelity and interpretability than MOB and CTree. CTree performed worst in these settings. This is often the case when there is a higher number of categorical features with low cardinality included in the dataset (which interact also with numeric features in the data set).
    \item use MOB and CTree in scenarios with smooth interactions (scenario linear smooth and linear mixed) for which these algorithms produce more stable trees while performing as well as SLIM and GUIDE. These settings are more likely for data sets with a high number of numerical features that are interacting with each other.
    However, smooth interactions can often only be modeled well by a large number of binary splits, which makes MBTs difficult to interpret on such data. Thus, depending on the underlying feature interactions, MBTs might not be the best choice. Global modeling approaches such as GA2M \cite{Lou.2013} or compboost \cite{Schalk.2018} should be considered or at least compared to MBTs in this case.
\end{enumerate}
If features can be separated into modeling and partitioning features (e.g., based on domain-specific knowledge), we recommend using MOB which has been developed and analyzed in detail for these settings and showed good fidelity and stable results \cite{Alemayehu.2018,Huber.2019}.

\subsection{Extensions Beyond Linearity}

When the main effects learned by the black box model are non-linear, the MBTs will not only split according to feature interactions but also according to non-linear main effects to approximate the main effects within regions by linear effects. This leads to deeper trees, which again are less interpretable. An alternative to fitting linear models within MBTs would be to use, for example, polynomial regression, splines, or GAMs. These models are able to account for non-linearity such that splits can be placed according to feature interactions. However, not all MBTs provide this flexibility to adapt the fitted model within the recursive partitioning algorithm (at least not out-of-the-box). \cite{Hu.2020} provide these alternatives, including efficient estimation procedures for SLIM. We apply SLIM with more flexible models fitted within the regions on a non-linear setting in Online~Appendix~\ref*{app:nonlinear}~\cite{HLDEC23} to demonstrate the differences and improved interpretability compared to the usage of linear models.
We leave adaptions and analyses of GUIDE, MOB, and CTree for these scenarios to future research.

It is also helpful to add a regularization term for settings with many potential noise features to obtain more interpretable and potentially more stable results. SLIM again allows adding any regularization term (e.g., Lasso regularization for feature selection) out of the box.
Such an analysis can be found in Online Appendix~\ref*{app:noise_feats}~\cite{HLDEC23} for the MBTs as surrogate models for the (correctly specified) linear model and as a standalone model on the DGP.
Further analyses using other ML models and diverse hyperparameters are also a matter of future research.

%% file: chapters/7_conclusion.tex
While SLIM and GUIDE performed strongly in most of our simulation settings, they often showed less stable results compared to MOB and CTree in our analyses.
In some scenarios, the tree size varied greatly for both algorithms. This observation might depend on the chosen hyperparameter configuration. Thus, SLIM and GUIDE could be improved by tuning the early stopping hyperparameters or by adding a post-pruning step to receive more stable results. 

Furthermore, SLIM might be sensitive regarding a selection bias which is common in recursive partitioning algorithms based on exhaustive search. In contrast, MOB, CTree, and GUIDE circumvent that problem by a two-step approach in their splitting procedure which is based on parameter stability tests \cite{Schlosser.2019}.
How the selection bias influences the trees fitted in these settings as well as the investigation of an extended setup of more complex scenarios and real-world settings are interesting open questions to analyze in future work.




In conclusion, it can be said that MBT algorithms are a promising addition -- although not a universal solution -- to interpreting the black box models by surrogate models. By combining decision rules and (non-linear) main effect models, we might achieve high fidelity as well as high interpretability at the same time. However, interpretability decreases very quickly with a high number of regions. Thus, the trade-off between fidelity and interpretability remains, and the compromise to be found depends on the underlying use case. 

%% file: chapters/appendix.tex
\section{Details on Quantifying the Rand index}
\label{app:sim_setup}

The detailed steps for the quantification of the RI in the simulations in Section \ref*{sec:experiments} can be described as follows:
\begin{enumerate}
    \item Simulate evaluation data ($50000$ observations) from the DGP
    \item \textbf{For each} simulation run in $1:100$ runs:
    \begin{enumerate}
        \item Simulate data ($n = 1500$) and perform train/test split ($\frac{2}{3}/\frac{1}{3}$)
        \item Train MBT on the training data, calculate fidelity measures on the train and test set, and extract the number of leaf nodes
        \item save the partitioning of the evaluation data defined through the trained MBT
    \end{enumerate}
    \item \textbf{For each} of the ($100(100-1)/2 = 4950$ MBT pairs
    \begin{enumerate}
        \item Sample 1000 observations from the evaluation data sets
        \item \textbf{If} both trees have the same number of leaf nodes, calculate the $RI$ for the two partitions of the sampled evaluation data subset
    \end{enumerate}
\end{enumerate}

\section{More Results and Details on Experiments}
Here, we provide more details on the experiments described in Section \ref*{sec:experiments}. Therefore, details on model configurations are described as well as further results and detailed analyses are provided.

\subsection{Hyperparameter Configurations}
\label{app:config}
In the experiments of Section \ref*{sec:experiments}, an XGBoost algorithm was used as a black box model (besides a correctly specified lm or GAM) with correctly specified interaction terms.
In Table \ref{tab:app_xgboost_config} the XGBoost hyperparameter configurations, which were used for the simulations of Section \ref*{sec:experiments}, are defined.
\begin{table}[!htb]
    \centering
     \caption{XGBoost hyperparameter configurations for the three scenarios linear smooth, linear categorical, and linear mixed of the experiments in Section \ref*{sec:experiments}.}
    \label{tab:app_xgboost_config}
    \begin{tabular}{l|r|r|r}
    \hline
    & linear smooth & linear categorical & linear mixed \\
    \hline
    max\_depth & 5 & 3 & 5 \\
    eta & 0.5 & 0.5 & 0.5 \\
    alpha & 1 & 0.5 & 2 \\
    gamma & 2 & 1 & 3.5 \\
    nrounds & 400 & 350 & 500\\
    \hline
    \end{tabular}
\end{table}

\subsection{Experimental Results: Linear Smooth}
\label{app:lin_smooth}
\begin{figure}[!htb]
\centering
    \centering
    \includegraphics[width=\textwidth]{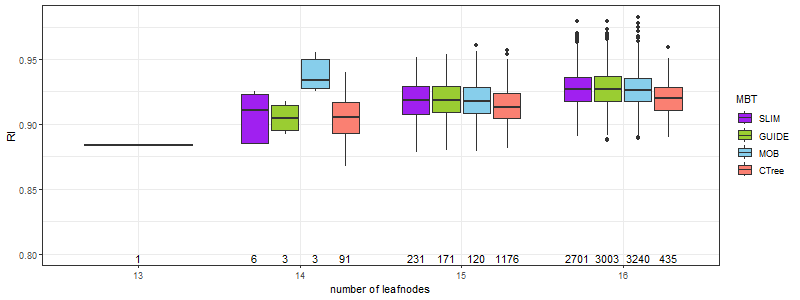}
    \caption{RI for the four MBT algorithms when used as a surrogate model for an lm model for the scenario linear smooth with $\alpha = \gamma = 0.05$. The numbers below the boxplots indicate the number of tree pairs, for which both trees have the respective number of leaf nodes.}
    \label{fig:ls_1000_lm_r2_train}
\end{figure}

 This section provides more detailed results on the scenario linear smooth described in Section \ref*{sec:experiments}.
A similar mean number of leaves for all MBTs are given by $\gamma = \alpha = 0.05$. Thus, the analysis of stability for the lm (Figure~\ref{fig:ls_1000_lm_r2_train}) is based on the results for these configurations to enable comparability between the different algorithms.

\subsection{Experimental Results: Linear Categorical}
\label{app:lin_cat}

This section provides more detailed results on the scenario linear categorical described in Section \ref*{sec:experiments}. Aggregated results on interpretability and fidelity are provided in Table \ref{tab:linear_abrupt_summary}.
The analysis of fidelity for the lm (Figure~\ref{fig:ls_1000_lm_r2_train}) is based on the configuration $\gamma = \alpha = 0.05$.
The results coincide with the results of the XGBoost model (Figure~\ref{fig:ls_1000_xgboost_r2_train}).
Hence, MOB and CTree lead to a worse fidelity and due to many more leaf nodes to a worse interpretability than SLIM and GUIDE for this scenario.

\begin{figure}[!htb]
\centering
    \includegraphics[width=\textwidth]{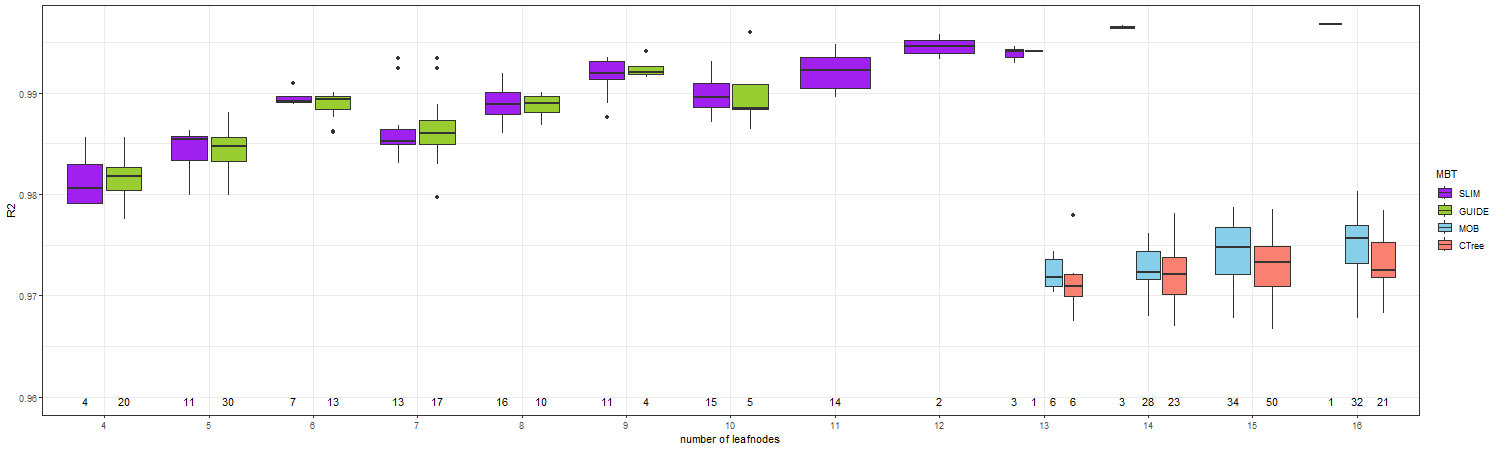}
    \caption{Test accuracy $R^2$ vs. number of leaf nodes for the four MBT algorithms as surrogate models for lm for scenario linear categorical with $n_{train}=1000$ and $n_{test} = 500$, $\alpha = \gamma = 0.05$. The numbers below the boxplots indicate the number of trees (from 100 simulation runs) which have the respective number of leaf nodes for the regarded algorithm.}
    \label{fig:la_1000_lm_r2_test}

\end{figure}

\begin{table}[!htb]

\centering
\caption{Simulation results on 100 simulation runs for all four MBTs on scenario linear categorical with sample sizes $n_{train}=1000$ and $n_{test} = 500$ for different values of $\gamma$ and $\alpha$. The mean (standard deviation) fidelity on the training data for the GAM is 0.9702 (0.0018) and for the XGBoost is 0.9876 (0.0015). On the test data set the respective fidelity values for the GAM are 0.9694 (0.0029) and for the XGBoost are 0.9778 (0.0031).}
\label{tab:linear_abrupt_summary}
\begin{tabular}[t]{l|l|r|r|r|r|r|r|r|r|r}
\hline
Black box& MBT&$\gamma$/$\alpha$ & \multicolumn{3}{c|}{Number of Leaves}&\multicolumn{2}{c|}{$R^2_{train}$}&\multicolumn{2}{c|}{$R^2_{test}$}  & Share\\
\hline
& &  & mean & min & max & mean & sd & mean & sd &  $x_2$\\
\hline
xgboost & SLIM & 0.15  & 2.00 & 2 & 2 & 0.8321 & 0.8321 & 0.8323 & 0.8323 & 0.0000\\
xgboost & GUIDE & 0.15  & 2.00 & 2 & 2 & 0.8321 & 0.8321 & 0.8323 & 0.8323 & 0.0000\\
xgboost & MOB  & 0.001 & 13.45 & 11 & 16 & 0.9793 & 0.9793 & 0.9729 & 0.9729 & 0.8865\\
xgboost & CTree & 0.001 & 11.96 & 10 & 14 & 0.9602 & 0.9602 & 0.9545 & 0.9545 & 0.9914\\
xgboost & SLIM & 0.10  & 4.00 & 4 & 4 & 0.9923 & 0.9923 & 0.9870 & 0.9870 & 0.0000\\
xgboost & GUIDE & 0.10 & 4.00 & 4 & 4 & 0.9923 & 0.9923 & 0.9870 & 0.9870 & 0.0000\\
xgboost & MOB & 0.010 & 14.38 & 13 & 16 & 0.9831 & 0.9831 & 0.9765 & 0.9765 & 0.8656\\
xgboost & CTree  & 0.010 & 12.76 & 10 & 15 & 0.9612 & 0.9612 & 0.9550 & 0.9550 & 0.9897\\
xgboost & SLIM & 0.05 & 4.00 & 4 & 4 & 0.9923 & 0.9923 & 0.9870 & 0.9870 & 0.0000\\
xgboost & GUIDE & 0.05 & 4.00 & 4 & 4 & 0.9923 & 0.9923 & 0.9870 & 0.9870 & 0.0000\\
xgboost & MOB & 0.050 & 14.63 & 13 & 16 & 0.9837 & 0.9837 & 0.9771 & 0.9771 & 0.8614\\
xgboost & CTree & 0.050 & 13.46 & 10 & 16 & 0.9623 & 0.9623 & 0.9558 & 0.9558 & 0.9838\\
\hline
gam & SLIM & 0.15 & 2.00 & 2 & 2 & 0.8528 & 0.8528 & 0.8513 & 0.8513 & 0.0000\\
gam & GUIDE & 0.15 & 2.00 & 2 & 2 & 0.8528 & 0.8528 & 0.8513 & 0.8513 & 0.0000\\
gam & MOB & 0.001 & 13.53 & 11 & 15 & 0.9773 & 0.9773 & 0.9718 & 0.9718 & 0.9824\\
gam & CTree & 0.001 & 13.89 & 11 & 16 & 0.9773 & 0.9773 & 0.9720 & 0.9720 & 0.9973\\
gam & SLIM & 0.10 & 2.64 & 2 & 4 & 0.8972 & 0.8972 & 0.8937 & 0.8937 & 0.0000\\
gam & GUIDE & 0.10 & 2.64 & 2 & 4 & 0.8972 & 0.8972 & 0.8937 & 0.8937 & 0.0000\\
gam & MOB & 0.010 & 14.28 & 13 & 16 & 0.9784 & 0.9784 & 0.9728 & 0.9728 & 0.9721\\
gam & CTree & 0.010 & 14.47 & 12 & 16 & 0.9779 & 0.9779 & 0.9725 & 0.9725 & 0.9949\\
gam & SLIM & 0.05 & 8.56 & 4 & 16 & 0.9910 & 0.9910 & 0.9893 & 0.9893 & 0.0017\\
gam & GUIDE & 0.05 & 6.06 & 4 & 13 & 0.9875 & 0.9875 & 0.9859 & 0.9859 & 0.0084\\
gam & MOB & 0.050 & 14.92 & 13 & 16 & 0.9797 & 0.9797 & 0.9740 & 0.9740 & 0.9572\\
gam & CTree & 0.050 & 14.86 & 13 & 16 & 0.9783 & 0.9783 & 0.9729 & 0.9729 & 0.9925\\

\hline
\end{tabular}

\end{table}

\subsection{Experimental Results: Linear Mixed}
\label{app:lin_mixed}

This section provides more detailed results on the scenario linear mixed described in Section \ref*{sec:experiments}. Aggregated results on interpretability and fidelity are provided for all four MBT algorithms in Table \ref{tab:linear_mixed_summary}.

For $\alpha = \gamma = 0.05$ the mean number of leaf nodes is similar for all four MBT algorithms. Thus, these configurations are chosen for the analyses in Section \ref*{sec:experiments} to enable comparability between the different MBT algorithms.
Table \ref{tab:linear_mixed_summary} shows a slightly higher fidelity for SLIM and GUIDE compared to MOB and CTree for this configuration.
\begin{table}[!htb]
\centering
\caption{Simulation results on 100 simulation runs for all four MBTs on scenario linear mixed with sample sizes $n_{train}=1000$ and $n_{test} = 500$ for different values of $\gamma$ and $\alpha$. The mean (standard deviation) fidelity on the training data for the lm is 0.9902 (0.0006) and for the XGBoost is 0.9859 (0.0014). On the test data set the respective fidelity values for the lm are 0.9898 (0.0008) and for the XGBoost are 0.9682 (0.0042). The column ``Share'' defines the relative number of partitioning steps which used the numeric features $\textbf{x}_1$ or $\textbf{x}_2$ for splitting.}
\label{tab:linear_mixed_summary}
\begin{tabular}[t]{l|l|r|r|r|r|r|r|r|r|r}
\hline
Black box& MBT&$\gamma$/$\alpha$ & \multicolumn{3}{c|}{Number of Leaves}&\multicolumn{2}{c|}{$R^2_{train}$}&\multicolumn{2}{c|}{$R^2_{test}$}  & Share\\
\hline
& &  & mean & min & max & mean & sd & mean & sd &  $x_1$ or $x_2$\\
\hline
xgboost & SLIM & 0.15 & 4.47 & 2 & 13 & 0.9067 & 0.0336 & 0.9013 & 0.0339 & 0.9486\\
xgboost & GUIDE & 0.15 & 4.37 & 2 & 13 & 0.9059 & 0.0335 & 0.9005 & 0.0339 & 0.9453\\
xgboost & MOB &  0.001 & 14.82 & 13 & 17 & 0.9853 & 0.0018 & 0.9745 & 0.0047 & 0.9735\\
xgboost & CTree  & 0.001 & 15.03 & 13 & 17 & 0.9850 & 0.0017 & 0.9743 & 0.0042 & 0.9949\\
xgboost & SLIM & 0.10 & 12.80 & 7 & 16 & 0.9832 & 0.0089 & 0.9724 & 0.0103 & 0.9044\\
xgboost & GUIDE & 0.10  & 12.48 & 6 & 16 & 0.9822 & 0.0098 & 0.9715 & 0.0112 & 0.8737\\
xgboost & MOB &  0.010 & 14.94 & 13 & 17 & 0.9854 & 0.0017 & 0.9746 & 0.0046 & 0.9727\\
xgboost & CTree  & 0.010 & 15.07 & 13 & 17 & 0.9850 & 0.0017 & 0.9743 & 0.0042 & 0.9947\\
xgboost & SLIM & 0.05 & 14.80 & 12 & 17 & 0.9870 & 0.0018 & 0.9764 & 0.0044 & 0.9068\\
xgboost & GUIDE & 0.05  & 14.47 & 12 & 17 & 0.9863 & 0.0022 & 0.9758 & 0.0047 & 0.8683\\
xgboost & MOB & 0.050 & 14.94 & 13 & 17 & 0.9854 & 0.0017 & 0.9746 & 0.0046 & 0.9727\\
xgboost & CTree & 0.050 & 15.07 & 13 & 17 & 0.9850 & 0.0017 & 0.9743 & 0.0042 & 0.9947\\
\hline
lm & SLIM & 0.15  & 3.20 & 2 & 13 & 0.8879 & 0.0309 & 0.8806 & 0.0331 & 0.9705\\
lm & GUIDE & 0.15  & 3.17 & 2 & 13 & 0.8872 & 0.0308 & 0.8799 & 0.0329 & 0.9707\\
lm & MOB &0.001 & 14.99 & 13 & 17 & 0.9882 & 0.0016 & 0.9838 & 0.0021 & 0.9637\\
lm & CTree & 0.001 & 15.05 & 13 & 17 & 0.9880 & 0.0016 & 0.9841 & 0.0019 & 0.9994\\
lm & SLIM & 0.10  & 13.07 & 5 & 16 & 0.9875 & 0.0098 & 0.9843 & 0.0108 & 0.8766\\
lm & GUIDE & 0.10  & 12.66 & 7 & 16 & 0.9866 & 0.0095 & 0.9834 & 0.0106 & 0.8676\\
lm & MOB &  0.010 & 14.99 & 13 & 17 & 0.9882 & 0.0016 & 0.9838 & 0.0021 & 0.9637\\
lm & CTree & 0.010 & 15.05 & 13 & 17 & 0.9880 & 0.0016 & 0.9841 & 0.0019 & 0.9994\\
lm & SLIM & 0.05 & 14.78 & 12 & 16 & 0.9913 & 0.0020 & 0.9885 & 0.0028 & 0.8723\\
lm & GUIDE & 0.05 & 14.38 & 12 & 16 & 0.9905 & 0.0022 & 0.9876 & 0.0029 & 0.8611\\
lm & MOB & 0.050 & 14.99 & 13 & 17 & 0.9882 & 0.0016 & 0.9838 & 0.0021 & 0.9637\\
lm & CTree & 0.050 & 15.05 & 13 & 17 & 0.9880 & 0.0016 & 0.9841 & 0.0019 & 0.9994\\
\hline

\end{tabular}

\end{table}

\subsection{Non-linear effects}
\label{app:nonlinear}
While the scenario definitions so far were only based on linear main effects, the following scenario will investigate how the choice of different objective functions affects the interpretability and fidelity of SLIM MBTs when non-linear main effects are included in the DGP.
Since \cite{Hu.2020} provide flexible options to efficiently estimate non-linear main effects in the leaf nodes, only the SLIM algorithm is used.

\subsubsection{Scenario definition}
Let $X_1,\ldots,X_5 \sim \mathcal{U}(-1,1)$, $\textbf{x}_6 \sim Bern(0.5)$ and the DGP based on $n$ realizations is defined by $y = f(\textbf{x}) + \epsilon$ with $ f(\textbf{x}) = \textbf{x}_1 + 2 \textbf{x}_2^2 + \textbf{x}_3log(abs(\textbf{x}_3)) + \textbf{x}_4\textbf{x}_5 + \textbf{x}_1\textbf{x}_4\mathbb{1}_{(\textbf{x}_6 = 0)}$ and $\epsilon \sim \mathcal{N}(0,  0.01 \cdot \sigma^2(f(x)))$.

SLIM is fitted with the following four different model types in the leaf nodes:
\begin{enumerate}
    \item Linear regression model (lm)
    \item Polynomial regression of degree $2$ with lasso penalization (penalized poly)
    \item Linear regression with unpenalized linear B-spline transformations of the features (B-splines)
    \item GAMs with integrated smoothness estimation \cite{Wood.2011}
\end{enumerate}

Since the models fitted in the leaf nodes vary in complexity and thus interpretability, it is not sufficient to consider solely the number of leaf nodes as a measure of interpretability. In addition, the following criteria are analyzed:
\begin{itemize}
    \item Effective degrees of freedoms of the leaf node models (lm and penalized poly), as sparse models are easier to interpret
    \item Proportion of splits for which features are chosen that are involved in feature interactions (in the DGP) vs. proportion of splits for which features are selected that only contain non-linear main effects.
    \item Interpretable formulas vs. only visual interpretability
\end{itemize}

In order to enable a comparison of the interpretability, the $R^2$ is used as an early stopping parameter in this scenario in addition to the early stopping procedure with $\gamma$. As soon as the $R^2$ exceeds a certain value in a node, no further split for this node is performed. We choose $\gamma = 0.05$ and an $R^2$ value of $0.9$ for as early stopping configurations for the described simulation setting.



We simulate $n =3000$ observations (2000 for training and 1000 testing). The SLIM MBTs are fitted as a surrogate model on the predictions of an XGBoost model. The hyperparameter configurations of the XGBoost model are defined in Table \ref{tab:app_xgboost_config_nonlinear}.

\begin{table}[!htb]
    \centering
    \caption{Tuned hyperparameter configurations for the XGBoost algorithm with correctly specified interaction effects for the scenario non-linear.}
\label{tab:app_xgboost_config_nonlinear}
    \begin{tabular}{l|r}
    \hline
    & non-linear  \\
    \hline
    max\_depth & 4 \\
    eta & 0.825 \\
    alpha & 0.75 \\
    gamma & 1 \\
    nrounds & 700 \\
    \hline
    \end{tabular}

\end{table}

\subsubsection{Results}
Table \ref{tab:linear_mixed_1_interpretability} provides an overview of the simulation results regarding the interpretability measures for the different model types used within the SLIM MBTs. For SLIM with linear regression models in the nodes, this results in large trees which provide low interpretability due to the high number of leaf nodes.
In addition, on average $61\%$ of all partitioning steps do not use a feature involved in an interaction for splitting, but split based on an insufficiently modeled main effect. Moreover, a high number of features is used for splitting which further reduces interpretability.
The advantage is that the models fitted in the leaf nodes provide an interpretable formula and not only a visual component to interpret the final results.

Using the second model type, meaning penalized polynomial regression, in the leave nodes when applying SLIM, leads to a comparable fidelity as using an lm but with fewer splits. Thus, interpretability increases in the sense that the number of leaf nodes drops. Also, the number of different split features is reduced, which again increases interpretability. While the proportion of partitioning steps that use main effect features for splitting is smaller, there is still more than one-third of splits that are not performed due to feature interactions.

Both, SLIM with linear B-Spline transformed features and GAMs require on average only two leaf nodes, i.e. one split, to achieve an $R^2$ of $0.9$. Thus, with regard to the number of leaf nodes the interpretability highly increases. However, the models in the leaf nodes can only be interpreted visually, and no interpretable formula is provided. Since the number of models (2) is very small, the degree of interpretability is comparatively high. Moreover, these models actually only split by interactions, as the non-linear main effects are already modeled sufficiently well.

If an interpretable formula is not explicitly required and a visual interpretation is sufficient, it is recommended to use flexible models such as splines in the leaf nodes. GAMs are preferable to unpenalized B-Splines in terms of their generalization error, however, it needs to be considered that it is computationally more expensive as shown in Table \ref{tab:performance_nonlinear}.

\begin{table}[!htb]
\centering
\caption{Simulation results with regard to interpretability for SLIM when different model types in the leaf nodes are fitted for the scenario non-linear. The different SLIM variants are applied as surrogate models on the XGBoost model. The interpretability measures are the number of leaf nodes, the number of split features used within the fitted trees, the share of splits which are based on main effect features, and the effective degrees of freedom.}
\label{tab:linear_mixed_1_interpretability}
\begin{tabular}[t]{l|l|r|r|r|r|r|r|r|r|r}
\hline
DGP/ black box & Model & \multicolumn{3}{c}{No. of leaves} & \multicolumn{3}{|c|}{No. of split feat} & Share & \multicolumn{2}{|c}{Df}  \\

\hline
& & mean & min & max & mean & min & max & main & mean & sd\\
\hline
XGBoost & lm & 19.82 & 2 & 33 & 4.76 & 1 & 6 & 0.6125 & 6.8671 & 0.1438\\
XGBoost & penalized poly & 3.58 & 2 & 7 & 2.08 & 1 & 4 & 0.3880 & 8.5499 & 1.1147\\
XGBoost & B-Splines & 1.86 & 1 & 3 & 0.84 & 0 & 1 & 0.0000 &  & \\
XGBoost & GAM & 1.90 & 1 & 4 & 0.88 & 0 & 2 & 0.0000 &  & \\

\hline
\end{tabular}

\end{table}

\begin{table}[htb]

\centering
\caption{Simulation results with regard to fidelity for SLIM when different model types in the leaf nodes are fitted for the scenario non-linear. The different SLIM variants are applied as surrogate models on the XGBoost model. Besides the $R^2$ values of the model fits, also the average required time in seconds to fit one tree is provided in the last column.}
\label{tab:performance_nonlinear}
\begin{tabular}[t]{l|l|r|r|r|r|r}
\hline
DGP/ black box & Model & \multicolumn{2}{c}{$R^2_{train}$} & \multicolumn{2}{|c|}{$R^2_{test}$} & time in sec \\
\hline
XGBoost & lm & 0.9200 & 0.0228 & 0.9023 & 0.0190 & 55.7771\\
XGBoost & penalized poly & 0.9213 & 0.0079 & 0.9150 & 0.0087 & 35.4093\\
XGBoost & B-Splines & 0.9382 & 0.0118 & 0.9296 & 0.0120 & 11.4670\\
XGBoost & GAM & 0.9348 & 0.0118 & 0.9289 & 0.0117 & 384.7403\\
\hline
XGBoost & XGBoost & 0.9386 & 0.0295 & 0.9199 & 0.0362 & 3.1163\\
\hline
\end{tabular}

\end{table}

\subsection{Linear smooth with noise features}
\label{app:noise_feats}
Here, we examine how noise features that have no influence on the target $y$ affect the MBTs. Therefore, we use the scenario linear smooth and we add six noise features to the underlying data set.
In addition to the four MBT algorithms used so far, SLIM models are fitted with lasso regularization \cite{Tibshirani.1996,Friedman.2010}. Lasso models allow to fit sparse models, i.e., a feature selection within the models fitted in each node is automatically included. The strength of the feature selection depends strongly on the penalization parameter. For all regularized SLIM models, the penalization parameter is selected using the BIC criterion \cite{Sabourin.2015}. However, in the case of $df = 3$ or $df = 2$, the additional restriction is defined so that the effective degrees of freedom (df) must not exceed this value. This enforces especially sparse models.

\subsubsection{Scenario definition}
Let $X_1, \ldots , X_{10} \sim \mathcal{U}(-1,1)$ where the DGP based on $n$ realizations is defined by $y = f(\textbf{x}) + \epsilon$ with
$ f(\textbf{x}) = \textbf{x}_1 + 4   \textbf{x}_2 + 3   \textbf{x}_2   \textbf{x}_3 $ and
$\epsilon \sim \mathcal{N}(0, 0.01 \cdot \sigma^2(f(\textbf{x})))$.
The MBTs are fitted as surrogates on lm predictions on a data set with sample size $ n = 3000$ ($2000$ training, $1000$ test observations) using the early stopping parameter configurations $\alpha = 0.001$ and $\gamma = 0.1$. The simulation is repeated $250$ times.

\subsubsection{Results}

The aim of the simulation is to investigate whether the noise features are incorrectly chosen as splitting features.
Table \ref{tab:linear_smooth_noisy_summary} shows an overview of the results of the described scenario.
Noise features are not used as splitting features if the MBTs are used as surrogates for an lm model.
However, when we applied the approaches directly on the data from the DGP (not on the model predictions) as a standalone model, SLIM and GUIDE use noise features for splitting. MOB and CTree always split with respect to non-noise features.
GUIDE shows the highest share of selecting noise features for splitting.
While the number of leaf nodes for SLIM with the lasso df 2 regularization decreases compared to using SLIM without regularization, the performance values remain on a comparable level and thus, the performance vs. interpretability trade-off improves when the respective regularization is applied in this scenario.

\begin{table}[htb]
\centering
\caption{
Simulation results on 250 simulation runs for an lm surrogate model and as a standalone (DGP) for all four MBTs on scenario linear mixed with noise features for different values of $\gamma$ and $\alpha$. The mean (standard deviation) fidelity on the training data for the lm is 0.9901 (0.0004). On the test data set the respective fidelity values for the lm are 0.9901 (0.0006). The column ``Share'' defines the proportion of trees in which at least one of the noise features is used for splitting. For comparison, we also show the results if the methods are directly applied to the DGP as a standalone model.}
\label{tab:linear_smooth_noisy_summary}
\begin{tabular}[t]{l|l|r|r|r|r|r|r|r|r}
\hline

Black box & MBT & Share  & \multicolumn{3}{c|}{number of leaves} & \multicolumn{2}{c|}{$R^2_{train}$} & \multicolumn{2}{c}{$R^2_{test}$}\\
\hline
& & $\textbf{x}_{noise}$ & mean & min & max & mean  & sd  & mean  & sd\\
\hline
lm & SLIM & 0.000  & 14.036 & 8 & 16 & 0.9987 & 0.0018 & 0.9984 & 0.0019\\
lm & SLIM lasso & 0.000  & 13.736 & 8 & 16 & 0.9985 & 0.0023 & 0.9982 & 0.0027\\
lm & SLIM lasso df 3 & 0.000  & 14.008 & 8 & 16 & 0.9985 & 0.0023 & 0.9983 & 0.0026\\
lm & SLIM lasso df 2 & 0.000  & 11.160 & 5 & 14 & 0.9979 & 0.0018 & 0.9977 & 0.0020\\
lm & GUIDE & 0.000  & 14.096 & 8 & 16 & 0.9988 & 0.0018 & 0.9984 & 0.0019\\
lm & MOB & 0.000  & 15.960 & 15 & 16 & 0.9995 & 0.0000 & 0.9993 & 0.0001\\
lm & CTree & 0.000  & 15.564 & 13 & 16 & 0.9994 & 0.0001 & 0.9992 & 0.0001\\
\hline
DGP & SLIM & 0.072 &  11.988 & 5 & 17 & 0.9880 & 0.0048 & 0.9854 & 0.0049\\
DGP & SLIM lasso & 0.092  & 11.048 & 5 & 16 & 0.9871 & 0.0049 & 0.9852 & 0.0051\\
DGP & SLIM lasso df 3 & 0.028  & 9.732 & 4 & 14 & 0.9863 & 0.0046 & 0.9848 & 0.0050\\
DGP & SLIM lasso df 2 & 0.028  & 9.648 & 4 & 15 & 0.9864 & 0.0044 & 0.9852 & 0.0047\\
DGP & GUIDE & 0.104  & 11.788 & 5 & 16 & 0.9880 & 0.0047 & 0.9854 & 0.0048\\
DGP & MOB & 0.000  & 11.096 & 8 & 14 & 0.9901 & 0.0005 & 0.9878 & 0.0007\\
DGP & CTree & 0.000  & 13.140 & 10 & 16 & 0.9904 & 0.0004 & 0.9882 & 0.0007\\
\hline
\end{tabular}

\end{table}